\documentclass[runningheads]{llncs}
\usepackage{graphicx}
\usepackage{subfigure}
\usepackage{comment}
\usepackage{amsmath,amssymb} 
\DeclareMathOperator{\sign}{sign}
\DeclareMathOperator{\vadvgen}{VAdvGen}
\DeclareMathOperator{\qadvgen}{QAdvGen}

\usepackage{color}
\usepackage[ruled]{algorithm2e}
\usepackage{booktabs}
\usepackage{multirow}
\usepackage{floatrow}
\usepackage{setspace}
\usepackage{hyperref}
\hypersetup{hidelinks,
    colorlinks=true,
    allcolors=black,
    linkcolor=black,
    filecolor=blue,      
    urlcolor=blue,
    citecolor=black,
}
\newfloatcommand{capbtabbox}{table}[][\FBwidth]

\begin{document}
\pagestyle{headings}
\mainmatter
\def\ECCVSubNumber{3245}  

\title{Semantic Equivalent Adversarial Data Augmentation for Visual Question Answering}

\titlerunning{Semantic Equivalent Adversarial Data Augmentation for VQA}
%
\author{Ruixue Tang\inst{1} \and
Chao Ma\inst{1}\thanks{Corresponding author.} \and
Wei Emma Zhang\inst{2} \and
Qi Wu\inst{2} \and
Xiaokang Yang\inst{1}}
\authorrunning{R. Tang, C. Ma, W. Zhang, Q. Wu, and X. Yang}
%
\institute{MoE Key Lab of Artificial Intelligence, AI Institute, Shanghai Jiao Tong University \\ \email{\{alicetang, chaoma, xkyang\}@sjtu.edu.cn} \and
University of Adelaide \\
\email{\{wei.e.zhang, qi.wu01\}@adelaide.edu.au}}
\maketitle

\begin{abstract}

Visual Question Answering (VQA) has achieved great success thanks to the fast development of deep neural networks (DNN). On the other hand, the data augmentation, as one of the major tricks for DNN, has been widely used in many computer vision tasks. 
However, there are few works studying the data augmentation problem for VQA and none of the existing image based augmentation schemes (such as rotation and flipping) can be directly applied to VQA due to its semantic structure -- an $\langle image, question, answer\rangle$ triplet needs to be maintained correctly. For example, a direction related Question-Answer (QA) pair may not be true if the associated image is rotated or flipped. In this paper, instead of directly manipulating images and questions, we use generated adversarial examples for both images and questions as the augmented data. The augmented examples do not change the visual properties presented in the image as well as the \textbf{semantic} meaning of the question, the correctness of the $\langle image, question, answer\rangle$ is thus still maintained. We then use adversarial learning to train a classic VQA model (BUTD) with our augmented data. We find that we not only improve the overall performance on VQAv2, but also can withstand adversarial attack effectively, compared to the baseline model. The source code is available at \href{https://github.com/zaynmi/seada-vqa}{https://github.com/zaynmi/seada-vqa}.

\end{abstract}

\section{Introduction}
Both computer vision and natural language processing (NLP) have made enormous progress on many problems using deep learning in recent years. Visual question answering (VQA) is a field of study that fuses computer vision and NLP to achieve these successes. The VQA algorithm aims to predict a correct answer to the given question referring to an image. The recent benchmark study \cite{vqafut} demonstrates that the performance of VQA algorithms hinges on the amount of training data. Existing algorithms can always benefit greatly from more training data. This suggests that data augmentation without manual annotations is an intuitive attempt to improve the VQA performance, just like its success on the other deep learning applications.

Existing Data augmentation approaches enlarge the training dataset size by either data warping or oversampling \cite{dasur}. Data warping transforms data and keeps their labels. Typical examples include geometric and color transformations, random erasing, adversarial training, and neural style transfer. Oversampling generates synthetic instances and adds them to the training set. Data augmentation has shown to be effective in alleviating the overfitting problem of DNNs \cite{dasur}.
However, data augmentation in VQA is barely studied due to the challenge of maintaining an $\langle image, question, answer\rangle$ triplet semantically correct. Neither geometric transform nor randomly erasing the image could preserve the answer. For example, when asking about \emph{What is the position of the computer?}, \emph{Is the car to the left or right of the trash can?}, flipping or rotating images results in the opposite answers. Randomly erasing the image associated with the question \emph{How many ...?} would miss the number of objects. Such transforms need tailored answers which are unavailable. On the textual side, it is challenging to come up with generalized rules for language transformation. Universal data augmentation techniques in NLP have not been thoroughly explored. Therefore, it is non-trivial to explore the data augmentation technique to facilitate VQA.

Previous works have generated reasonable questions based on the image content \cite{othervqg} and the given answer \cite{dualvqg}, namely Visual Question Generation (VQG). However, a significant portion of the generated questions either have grammatical errors or are oddly phrased.
In addition, they learn from the questions and images in the same target dataset, thus the generated data are drawn from the same distribution of the original data. Since the training and test data usually do not share the same distribution, the generated data could not help to relieve the overfitting.

In this paper, we propose to generate semantic equivalent adversarial examples of both visual and textual data as augmented data. Adversarial examples are strategically modified samples that could successfully fool the deep models to make incorrect predictions. However, the modification is imperceptible that keeps the semantics of data while driving the underlying distribution of adversarial examples away from that of the original data \cite{advprop}.  
In our method, visual adversarial examples are generated by an un-targeted gradient-based attacker \cite{scale}, and textual adversarial examples are paraphrases that could fool the VQA model (predicting a wrong answer) while keeping the questions semantically equivalent.  
The existence of adversarial examples not only reveals the limited generalization ability of ConvNets, but also poses security threats on the real-world deployment of these models. 

We adversarially train the strong baseline method Bottom-Up-Attention and Top-Down (BUTD) \cite{butd} on VQAv2 dataset \cite{vqa} with clean examples and adversarial examples generated on-the-fly. We regard the adversarial training as a regularizer acting in a period of training time.
Experimental results demonstrate that our proposed adversarial training framework not only better boosts the model performance on clean examples than other data augmentation techniques, but also improves the model robustness against adversarial attacks. 
Although there are few works studying the data augmentation problem for VQA \cite{vqada,cyc,ctm,cau}, they merely generate either new questions or images. To our best knowledge, our work is the first to augment both visual and textual data in VQA.

To summarize, our major contributions are threefold:
\begin{itemize}
    \item We propose to generate visual and textual adversarial examples to augment the VQA dataset. Our generated data preserve the semantics and explore the learned decision boundary to help improve the model generalization.
    \item We propose an adversarial training scheme that enables VQA models to take advantage of the regularization power of adversarial examples.
    \item We show that the model trained with our method achieves 65.16\% accuracy on the clean validation set, beating its vanilla training counterpart by 1.84\%. Moreover, the adversarially trained model significantly increases accuracy on adversarial examples by 21.55\%.
\end{itemize}

\section{Related Work}
\subsubsection{VQA.} 
A large number of VQA algorithms have been proposed, including 
spatial attention \cite{butd,stacked,hierarchical,mutan}, compositional approaches \cite{nmn,compose,e2e}, and bilinear pooling schemes \cite{multimodal,hadamard}. Spatial attention \cite{butd} is one of the most widely used methods for both natural and synthetic image VQA. A large portion of prior arts \cite{ban,count,graph,intra} are built upon the bottom-up top-down (BUTD) attention method \cite{butd}. We also choose the BUTD as our baseline VQA model. Instead of developing a more sophisticated answering machine, we propose a VQA data augmentation technique that can potentially benefit existing VQA methods since data is the fuel.

\subsubsection{Data Augmentation.} 
Compared to vision, a few efforts have been done on augmenting text for classification problems. 
Wei \emph{et al.} \cite{eda} make a comprehensive extension for text editing techniques on NLP data augmentation
and achieve gains on text classification. 
However, our study shows that it could degrade the model performance on the VQA task (see Section \ref{ex}). Other works generate paraphrases \cite{qanet,para} and add noise to smooth text data \cite{noise}. 
There are fewer works \cite{vqada,ctm,cyc,cau,patro2018differential} that learn data augmentation for VQA.
 Kafle \emph{et al.}\cite{vqada} do a pioneer work where they generate new questions by using semantic annotations on images. 
 Work of \cite{ctm} automatically generates entailed questions for a source QA pair 
, but it uses additional data in Visual Genome \cite{vg} to add diversity to the generated questions.
 Work of \cite{cyc} proposes a cyclic-consistent training scheme where it generates different rephrasings of question 
 and train the model such that the predicted answers across the generated and original questions remain consistent. 
The method \cite{cau} 
employ a GAN-based re-synthesis technique to automatically remove objects to strengthen the model robustness against semantic visual variations.
Note that all of these methods augment data in a single modality (text-only or image-only) and heavily rely on complex modules to achieve slight performance gains. 

\subsubsection{Adversarial Attack and Defense.} 
In recent years, research works \cite{szegedy,fgsm} add imperceptible perturbations to input images, named adversarial examples, to evaluate the robustness of deep neural networks against such perturbation attacks. In the NLP community, state-of-the-art textual DNN attackers \cite{belinkov2017synthetic,blohm2018comparing,ebrahimi2018adversarial} use a different approach from those in the visual community to generate textual adversarial examples. Our work is inspired by SCPNs \cite{scpn} and SEA \cite{sea} which generate paraphrases of the sentence as textual adversarial examples.
Meanwhile, previous works \cite{fgsm} show that training with adversarial examples can improve the model generalization on small dataset (e.g., MNIST),
but degrade the performance on large datasets (e.g., ImageNet), in the fully-supervised setting. 
Recent notable work \cite{advprop} suggests that adversarial training could boost model performance even on ImageNet with a well-designed training scheme. 
A number of methods \cite{att,fool} have investigated adversarial attack on the VQA task. However, they merely attack the image and do not discuss how the adversarial examples can benefit the VQA model.
To summarize, how adversarial examples can facilitate VQA remains an open problem. This work sheds light on utilizing adversarial examples as augmented data for VQA. 

\begin{figure}[t]
\centering
\includegraphics[width=4in]{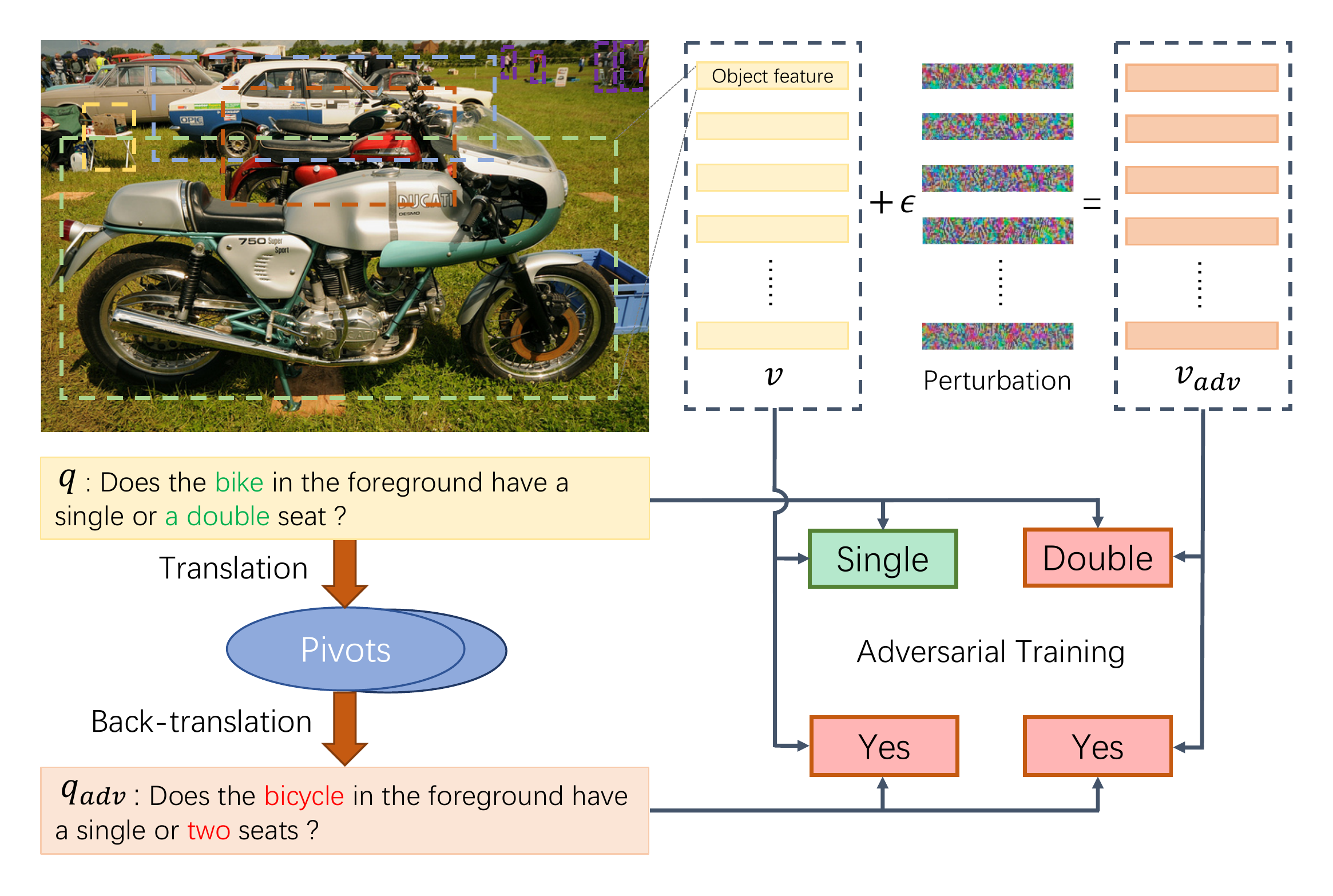}
\caption{Framework of the proposed data augmentation method. 
We generate adversarial examples of both visual and textual data as augmented data, which are passed through the VQA model to obtain incorrect answers.
The augmented and original data are jointly trained using the proposed adversarial training scheme, which can boost model performance on clean data while improving model robustness against attack.}
\label{overview}
\end{figure}

\section{Method}


We now introduce our data augmentation method to train a robust VQA model. 
As illustrated in Fig. \ref{overview}, given an $\langle image, question, answer \rangle$ triplet, we first generate the paraphrases of questions and store them, then, generate visual adversarial examples on-the-fly to obtain semantically equivalent additional training triplets, which are used in the proposed adversarial training scheme.
We describe them in detail as follows.

\subsection{VQA Model}
Answering questions about images can be formulated as the problem of predicting an answer \emph{a} given an image \emph{v} and a question \emph{q} according to a parametric probability measure:
\begin{align}
    \hat{a}=\arg\max_{a\in \mathcal{A}}p(a|v, q; \theta)
\end{align}
where $\theta$ represents a vector of all parameters to learn and $\mathcal{A}$ is a set of all answers. 
VQA requires solving several tasks at once involving both visual and textual inputs.
Here we use Bottom-Up-Attention and Top-Down (BUTD) \cite{butd} as our backbone model because it has become a golden baseline in VQA. In BUTD, region-specific image features extracted by fine-tuned Faster R-CNN \cite{frcnn} are utilized as visual inputs. In this paper, let 
$v = \left\{\overrightarrow{v_1},\overrightarrow{v_2},...,\overrightarrow{v_K}\right\}$ be a collection of visual features extracted from \emph{K} image regions and the question is a sequence of words $q=\left\{q_1,q_2,...,q_n\right\}$.
The $\langle image, question, answer\rangle$ triplet has a strong semantic relation that neither image nor question can be easily transformed to augment the training data while preserving original content. 

\subsection{Data Augmentation}
\label{dataaug}
Due to the risk of affecting answers, we avoid manipulating the raw inputs (i.e., images and questions) directly, such as cropping the image or changing the word order. Inspired by the adversarial attack and defense, we propose to generate adversarial examples as additional training data. In this section, we present how to generate adversarial examples of images and questions while preserving the original labels and how to use them to augment the training data.

\subsubsection{Visual Adversarial Examples Generation.}
Adversarial attacks are originated from the computer vision community. In general, the overarching goal is to add the least amount of perturbation to the input data to cause the desired misclassification. We employ an efficient gradient-based attacker Iterative Fast Gradient Sign Method (IFGSM)\cite{ifgsm} to generate visual adversarial examples. Before illustrating IFGSM, we firstly introduce FGSM, as IFGSM is an extension of it. 
Goodfellow \emph{et al.}\cite{fgsm} proposed the FGSM as a simple way to generate adversarial examples. We could apply it on visual input as: 
\begin{align}
\label{fgsm}
    v_{adv} = v + \epsilon \sign (\nabla_vL(\theta,v,q,a_{true}))
\end{align}
 where $v^{adv}$ is the adversarial example of $v$, $\theta$ is the set of model parameters, $L(\theta,v,q,a_{true})$ denotes the cost function used to train the VQA model, 
$\epsilon$ is the size of the adversarial perturbation. The attack backpropagates the gradient to the input visual feature to calculate $\nabla_vL(\theta,v,q,a_{true})$ while fixing the network parameters. Then, it adjusts the input by a small step  in the direction (i.e. $\sign (\nabla_vL(\theta,v,q,a_{true}))$) that maximize the loss. The resulting perturbed, $v_{adv}$, is then misclassified by the VQA model (e.g., the model predicts \emph{Double} in Fig. \ref{overview}). 
 
 A straightforward extension of FGSM is to apply it multiple times with small step size, referred to IFGSM as:
\begin{align}
\label{ifgsm}
    v_{adv}^0 = v, \quad  v_{adv}^{N+1}=Clip_{v,\epsilon}\left\{v_{adv}^N +\alpha \sign(\nabla_vL(\theta,v_{adv}^N,q,a_{true}))\right\}
\end{align}
where $Clip_{v,\epsilon}(A)$ denotes element-wise clipping $A$, with $A_{i,j}$ clipped to the range $[v_{i,j}-\epsilon,v_{i,j} +\epsilon]$, $\alpha$ is step size in each iteration. In this paper, we summarize gradient-based method as $\vadvgen(v,q)$. 

One-step methods of adversarial example generation generate a candidate adversarial image after computing only one gradient. 
Iterative methods apply many gradient updates. They typically do not rely on any approximation of the model and typically produce more harmful adversarial examples when running for more iterations. 
Our experimental results show that the accuracy of the BUTD vanilla trained model on visual adversarial examples generated by IFGSM is about 17\%$-$30\% for $\epsilon \in[0.3, 1.3]$.
It implies that adversarial examples have different distribution to normal examples.
\vspace{-10pt}
\subsubsection{Semantic Equivalent Questions Generation.}

To generate adversarial example $q_{adv}$ of a question, we cannot directly apply approaches from image DNN attackers since textual data is discrete. In addition, the perturbation size that measured by $L_p$ norm in image is also inapplicable for textual data. Moreover, the small changes in texts, e.g., character or word change, would easily destroy the grammar and semantics, rendering the possibility of attack failure. Adhere to the principle of not changing the semantics of input data, inspired by \cite{scpn,para}, we generate semantically equivalent adversarial questions by using a sequence-to-sequence paraphrasing model.


Here we propose a paraphrasing model based purely on neural networks and it is an extension of the basic encoder-decoder Neural Machine Translation (NMT) framework. In the neural encoder-decoder framework, the encoder (RNN) is used to compress the meaning of the source sentence into a sequence of vectors. The decoder, a conditional RNN language model, generates a target sentence word-by-word. The encoder takes a sequence of original question words $X=\left\{x_1,...,x_{T_x}\right\}$ as inputs, and produces a sequence of context. The decoder produces, given the source sentence, a probability distribution over the target sentence $Y=\left\{y_1,...,y_{T_y} \right\}$ with a softmax function:
\begin{align}
P(Y|X)=\prod_{t=1}^{T_y}P(y_t|y_{<t}, X)
\end{align}

 However, in the case of paraphrasing, there is not a path from English to English, but a path from English to a pivot language to English can be used. For example, the source English sentence $E_1$, is translated into a single French sentence $F$. Next, $F$ is translated back into English, giving a probability distribution over English sentences, $E_2$, which acts as paraphrase distribution:
 \begin{align}
P(E_2|E_1,F)=P(E_2|F)
\end{align}

Our paraphrasing model pivots through the set of $K$-best translations $\mathcal{F}=\left\{F_1,...,F_k\right\}$ of $E_1$. This ensures that multiple aspects (semantic and syntactic) of the source sentence are captured. Translating multiple pivot sentences into one sentence producing a probability distribution over the target vocabulary could be formed as:
\begin{align}
\label{mp}
P(y_t=w|y_{<t}, \mathcal{F})=\sum_{i=1}^{K}P(\mathcal{F}_i|E_1)\cdot P(y_t=w|y_{<t}, \mathcal{F}_i)
\end{align}

We further expand on the multi pivot approach by pivoting over multiple sentences in multiple languages (e.g. Portuguese). Deriving from Eq. \ref{mp}, we obtain $P(y_t=w|y_{<t}, \mathcal{F}^{Fr})$ and $P(y_t=w|y_{<t}, \mathcal{F}^{Po})$. 
Then averaging these two distributions, producing a multi-sentence, multi-lingual paraphrase probability:
\begin{align}
P(y_t=w|y_{<t}, \mathcal{F}^{Fr}, \mathcal{F}^{Po})=\frac{1}{2}(P(y_t=w|y_{<t}, \mathcal{F}^{Fr})+P(y_t=w|y_{<t}, \mathcal{F}^{Po}))
\end{align}
which is used to obtain the probability distributions over sentences:
\begin{align}
\label{pa}
P(E2|E1)=\prod_{t=1}^{T_{E_2}}P(y_t|y_{<t}, \mathcal{F}^{Fr}, \mathcal{F}^{Po})
\end{align}

We employ the pre-trained NMT model\footnote{\href{https://github.com/OpenNMT/OpenNMT-py}{https://github.com/OpenNMT/OpenNMT-py}} which is trained for English$\leftrightarrow$Portu-guese and English$\leftrightarrow$French to generate paraphrase candidates. 
A score \cite{sea} that measures the semantic similarity between paraphrase and its original text is defined as:
\begin{align}
\label{sscore}
S(q,q')=\min \left(1, \frac{P(q'|q)}{P(q|q)}\right)
\end{align}
where $P(q'|q)$ is the probability of a paraphrase $q'$ given original question $q$ defined in Eq. \ref{pa}, $P(q|q)$, which approximates how difficult it is to recover $q$, is used to normalize different distributions. We penalize those candidates with edit distance more than $e$ or unknown words by adding a large negative number $\lambda$ to the similarity score.
We select the paraphrase candidates with the top-k semantic scores as our $q_{adv}$.
The generation algorithm of  $q_{adv}$ is denoted $q_{adv}=\qadvgen(q)$. 
\begin{figure}[t]
\centering
\includegraphics[width=4in]{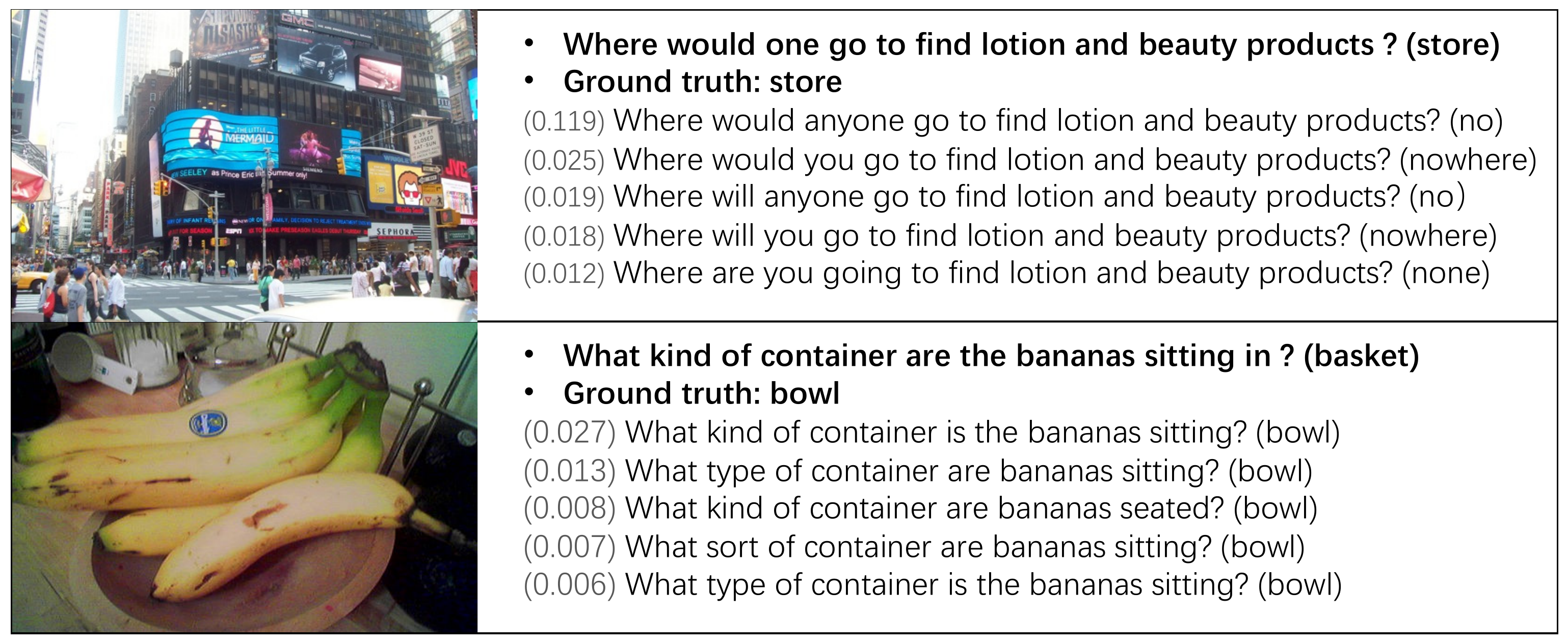}

\caption{Examples of our generated $q_{adv}$. The first question in bold in each block is the original question. The words in brackets are model predictions of the corresponding question; the numbers in brackets are the semantic score of $q_{adv}$.}
\label{seafig}
\end{figure}

Our paraphrases edit at least words to maintain syntax and semantics rather than exploring the linguistic variations regardless of the possibility of being perceived. 
We illustrate two examples of our $q_{adv}$ in Fig. \ref{seafig}.
They show that generated paraphrases could easily ``break'' the BUTD model.
A predicted label is considered “flipped” if it differs from the prediction on the corresponding original question (assume that we do not attack visual data in this part). We observe that $q_{adv}$ not only flip from positive predictions to negative ones but also correct the negative predictions to positive ones in some cases. Surprisingly, the flip rate of the vanilla trained model is 36.72\% causing an absolute accuracy drop of 10\%.
It suggests that there is brittleness in the model decision and indicates that the model exploits spurious correlations while making their predictions. 

\vspace{-8pt}
\subsection{Adversarial Training with Augmented Examples}  
Considering the adversarial training framework \cite{scale,advprop},  we treat adversarial examples as additional training samples and train networks with a mixture of adversarial and clean examples. The augmented questions are model-agnostic and generated before training, while visual adversarial examples are continually generated at every step of training. There are two kinds of visual adversarial examples depending on the question inputs:
\begin{align}
    v_{qc} = \vadvgen(v, q), \quad
    v_{qadv} = \vadvgen(v, q_{adv})
\end{align}
For each $(v,q)$ pair, we have 4 additional training pairs, $(v_{qc},q)$, $(v_{qadv},q)$, $(v_{qc},q_{adv})$ and $(v_{qadv},q_{adv})$. All these four pairs are semantically equivalent which means they hold the same ground truth answer. We maintain the original $\langle image, question, answer \rangle$ triplet but augment the original example at least four times, in the case of only one $q_{adv}$ generated.
We formulate a loss function that allows control of the relative weight of additional pairs in each batch:
\begin{align}
\label{loss}
    Loss \notag&=L(\theta,v, q,a_{true})+w\Big(L(\theta,v_{qc}, q,a_{true}) +L(\theta,v_{qadv},q,a_{true})\\
    &+L(\theta,v_{qc}, q_{adv},a_{true})+L(\theta,v_{qadv}, q_{adv},a_{true})\Big)
\end{align}
where $L(\theta,v,q,a_{true})$ is a loss on a batch of $v$ and $q$ examples with true answer $a_{true}$, $w$ is a parameter which controls the relative weight of adversarial examples in the loss. Our main goal is to improve network performance on clean images by leveraging the regularization power of adversarial examples. We empirically find that training with a mixture of adversarial and clean examples from beginning to end would not converge well on clean samples. Therefore, we mix them in a period of training time and fine-tune with clean examples in the rest epochs. Not only does this boost the performance on clean examples, but also improves the robustness of the model to adversarial examples.
We present our adversarial training scheme in Algorithm \ref{al}.
\setlength {\belowcaptionskip} {10pt}
\begin{algorithm}[t]
\label{al}
\caption{Pseudo code of our adversarial training}
\LinesNumbered 
\KwIn{A set of clean visual and textual examples $v$, $q$ with answers $a$ }
\KwOut{Network parameter $\theta$}
$q_{adv}=\qadvgen(q)$\;  
\For{each training step i}
{Sample a mini-batch of clean visual examples $v^b$, clean textual examples $q^b$ and textual adversarial examples $q^b_{adv}$ with answer $a^b$\;
\eIf{i is in adversarial training period time}
{Generating the corresponding mini-batch of additional training pairs $(v^b_{qc},q^b)$, $(v^b_{qadv},q^b)$, $(v^b_{qc},q^b_{adv})$ and $(v^b_{qadv},q^b_{adv})$\;
Minimize the loss in Eq. \ref{loss} w.r.t. network parameter $\theta$}
{
Minimize the loss $L(\theta,v^b,q^b,a^b)$ w.r.t. network parameter $\theta$
}
}
\Return $\theta$
\end{algorithm}
\vspace{-10pt}
\section{Experiments}
\label{ex}
\subsection{Experiments Setup}
\subsubsection{Dataset.}
We conduct experiments on the VQAv2 \cite{vqa}, which is improved from the previous version to emphasize visual understanding by reducing the answer bias in the dataset. VQAv2 contains 443K train, 214K validation and 453K test examples. The annotations for the test set are unavailable except for the remote evaluation servers. 
We provide our results on both validation and test set, and perform ablation study on the validation set.
\vspace{-10pt}
\subsubsection{VQA Architectures.}
We use a strong baseline Bottom-Up-Attention and Top-Down (BUTD) \cite{butd} which combines a bottom-up and a top-down attention mechanism to perform VQA, with the bottom-up mechanism generating object proposals from Faster R-CNN \cite{frcnn}, and the top-down mechanism predicting an attention distribution over those proposals. This model obtained first place in the 2017 VQA Challenge. Following setting in \cite{butd,tip}, we use a maximum of 100 object proposals per image, which are 2048 dimensional features, as visual input. We represent question words as 300 dimensional embeddings initialized with pre-trained GloVe vectors \cite{glove}, and process them with a one-layer GRU to obtain a 1024 dimensional question embedding. 
\vspace{-10pt}
\subsubsection{Training Details.}
For fair comparison, we train the BUTD baseline and our framework using Adamax \cite{adam} with a batch size of 256 on the training split for a total of 25 epochs.
Baseline achieves 63.32\% accuracy on the validation set and we save this checkpoint to evaluate the attackers in the following. To augment data by our framework, we adjust the learning rate at different stages. 
We set an initial learning rate of 0.001, and then decay it after five epochs at the rate of 0.25 for every two epochs. 
We inject the additional data merely in a period of epochs $(start, end)$, where $start$ is the epoch when we start adversarial training and $end$ is the epoch when we start standard training. We set the number of iterations $n$ of IFGSM to 2 and the number of generated paraphrases per question to 1 for saving training time. In paraphrase generating, we set the edit distance threshold $e=4$ and penalization score $\lambda=-10$.
To avoid \emph{label leaking} effect \cite{scale}, we replace the true label in Eq. \ref{fgsm} and \ref{ifgsm} with the most likely label predicted by the model when adversarial training. Our best result is achieved by using values $\epsilon=0.3, \alpha=0.0625, w=50$. These hyperparameters are chosen based on grid search, and other settings are tested in the ablation studies.
\subsection{Results}
\subsubsection{Overall Performance.}
Table \ref{tab1} shows the results on VQAv2 validation, test-dev and test-std sets. We compare our method with the BUTD vanilla training setting. 
Our method outperforms vanilla trained baseline, making gains of 1.82\%, 2.55\%, 2.6\% on validation, test-dev and test-std set, respectively. Furthermore, our adversarial training method only consumes a small amount of additional time (4 min for an epoch) while allows for a considerable increase in standard accuracy.

\vspace{-12pt}
\subsubsection{Comparison with Other Data Augmentation Methods.}
We compare our method with related VQA data augmentation method CC \cite{cyc}, and NLP data augmentation method EDA \cite{eda} and report the results on VQAv2 in Table \ref{tab1}.
The model of CC is trained to predict the same (correct) answer for a question and its rephrasing, which are generated by a VQG module in their training scheme. Their outperforming validation accuracy is in contrast to the less competitive accuracy on the test-dev set. It reveals CC is less capable of generalizing on unseen data. Other related studies (e.g., CausalVQA \cite{cau}, CTM\cite{ctm}) explore VQA data augmentation as a complementary result for constructing a new VQA dataset,  and they evaluate their data augmentation method on the new dataset instead of VQAv2, so it is hard to compare our method with them.
EDA is a text editing technique boosting model performance on the text classification task. We implement it to generate three augmented questions per original question and set the percent of words in a question that are changed $\alpha=0.1$. However, results (see Table \ref{tab1}) show that EDA could degrade the performance on clean data and make a 0.59\% accuracy drop. It demonstrates that text editing techniques for generating question are not applicable as large numbers of questions are too short that could not be allowed to insert, delete or swap words. Moreover, sometimes the text editing may change the original semantic meaning of the question, which leads to noisy and even incorrect data.

Since our augmented data might be regarded as injecting noise to original data, we also set comparison by injecting random noise with a standard deviation of 0.3 (same as our $\epsilon$ in reported results) to visual data. Random noise, as well, could be regarded as a naive attacker that causes a 0.9\% absolute accuracy drop on the vanilla model. However, jointly training with clean and noising data could not boost the performance on clean data, as reported in Table \ref{tab1}. It proves that our generated data are drawn from the proper distribution that let the model take full advantage of the regularization power of adversarial examples.
\setlength{\tabcolsep}{4.5pt}
\begin{table}[t]

\begin{center}
\caption{Performance and ablation studies on VQAv2.0. All models listed here are single model, which trained on the training set to report \emph{Val} scores and trained on training and validation sets to report \emph{Test-dev} and \emph{Test-std} scores. The first row represents the vanilla trained baseline model. The rows begin with $+$ represents the data augmentation method added to the first row. EDA-3 represents that we generate three augmented questions per original questions using EDA \cite{eda}. $\dagger$ This method is implemented based on a stronger BUTD (see \cite{cyc}) and obtains a relatively small improvement (0.48\%) on validation score, even so, its test-dev score is surpassed by our method.}
\label{tab1}
\begin{tabular}{lcccccc}
\toprule
\multirow{2}{*}{Method} & \multirow{2}{*}{Val} & \multicolumn{4}{c}{Test-dev} & \multirow{2}{*}{Test-std}  \\
\cmidrule(r){3-6} 
& & Overall    &  Yes/no  &   Number & Others \\
\midrule
BUTD \cite{butd} &63.32  & 65.23  & 81.82  & 44.21 & 56.05 & 65.67 \\
\ \ +Noise            & 63.28 & 64.80 & 81.03 & 43.96 & 55.70 & -\\
\ \ +EDA-3 \cite{eda} & 62.73 &-  &-  & - & - & -\\
\ \ +CC \cite{cyc}$\dagger$ & 65.53 & 67.55 &-  & - &-  & -\\
\hline
\ \ +Ours               & $\mathbf{65.16}$ & $\mathbf{67.78}$  & $\mathbf{84.08}$ & $\mathbf{47.55}$ & $\mathbf{58.48}$ &$\mathbf{68.27}$\\
\ \ +Ours $w/o$ Aug-Q  & 65.05 &67.58  & 83.85 & 47.34 & 58.31 & -\\
\ \ +Ours $w/o$ Aug-V & 64.69  & 67.45 & 83.55 & 46.96 & 58.37 & -\\

\bottomrule
\end{tabular}
\end{center}
\vspace{-3mm}
\end{table}
\setlength{\tabcolsep}{1.8pt}

\subsection{Analysis}
\subsubsection{Training Set Size Impact.}
Furthermore, we conduct experiments using a fraction of the available data in the training set. As overfitting tends to be more severe when training on smaller datasets, we show that our method has more significant improvements for smaller training sets. We run both vanilla training and our method for the following training set fractions (\%): $\left\{20, 40, 60, 80\right\}$. Performances are shown in Table \ref{size}. The best accuracy without augmentation, 63.32\%, was achieved using 100\% of the training data. Our method surpasses it with 80\% of the training data, achieving 64.27\%.
\begin{table*}[t]
\begin{floatrow}
\capbtabbox{
 \setlength{\tabcolsep}{4pt}
 \begin{tabular}{ccc}
\toprule
 Training set size & BUTD & +Ours \\
 \midrule
 80\% & 62.77 & 64.27 \footnotesize{(+1.50)}\\
 60\% & 61.55 & 63.11 \footnotesize{(+1.56)}\\
 40\% & 59.47 & 61.35 \footnotesize{(+1.88)}\\
 20\% & 55.45 & 57.39 \footnotesize{($\mathbf{+1.94}$)}\\
 \bottomrule
 \end{tabular}
}{
 \caption{Validation accuracy (\%) across BUTD with and without our framework on different training set sizes.}
 \label{size}
 }
\capbtabbox{
\setlength{\tabcolsep}{4pt}
 \begin{tabular}{cc}
 \toprule
 $(start, end)$ & Accuracy \\
 \midrule
 (5,25) & 63.93 \\
 (10,25) & 64.08\\
 (10,15) &$\mathbf{65.16}$\\
 (15,20) & 64.18\\
 \bottomrule
 \end{tabular}
}{
 \caption{Validation accuracy (\%) of our method using different adversarial training period. }
 \label{epoch}
}
\end{floatrow}
\vspace{-2mm}
\end{table*}
\begin{figure}[t]
\centering
\subfigure[]{
\centering
\includegraphics[width=1.2in]{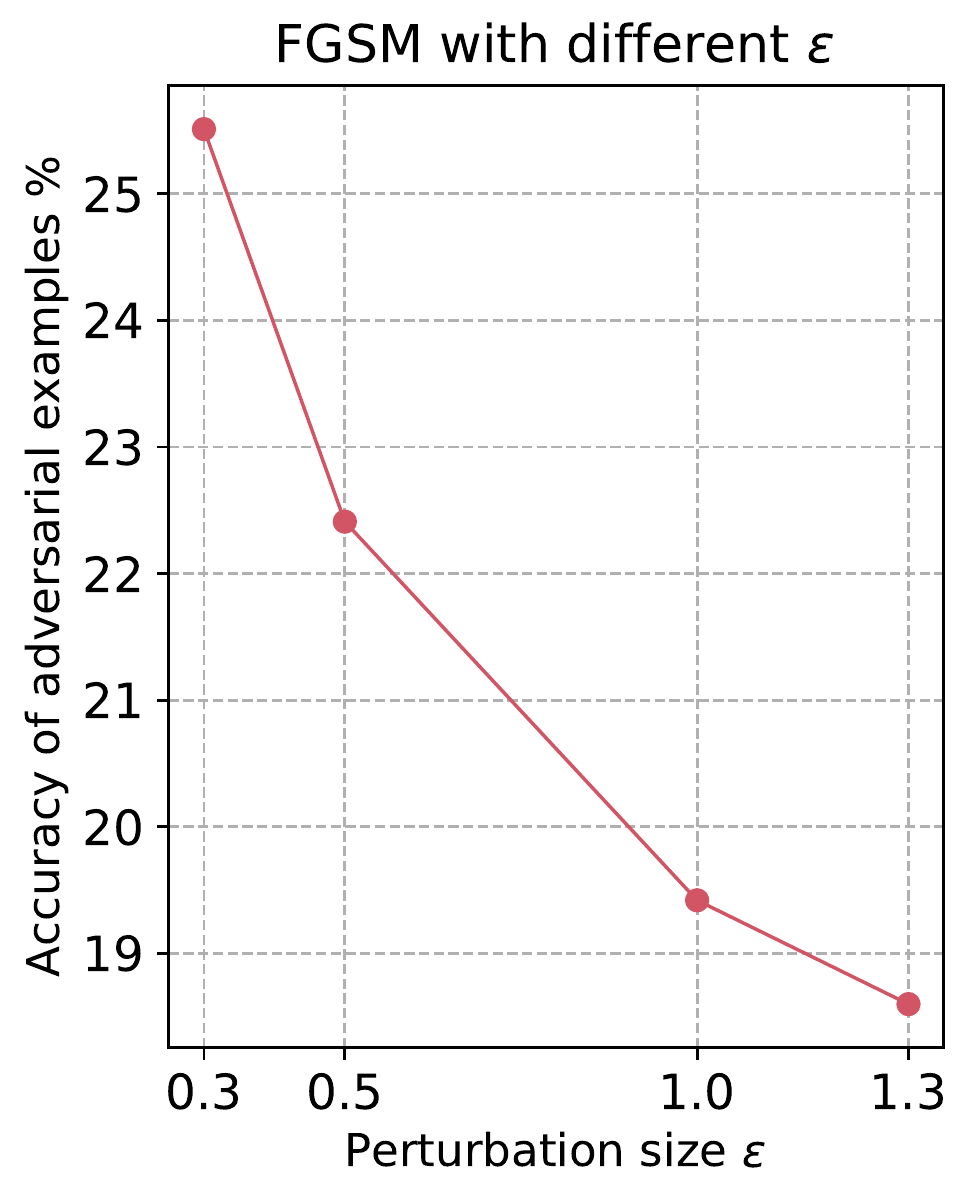}\label{fgsme}
}%
\subfigure[]{
\centering
\includegraphics[width=1.2in]{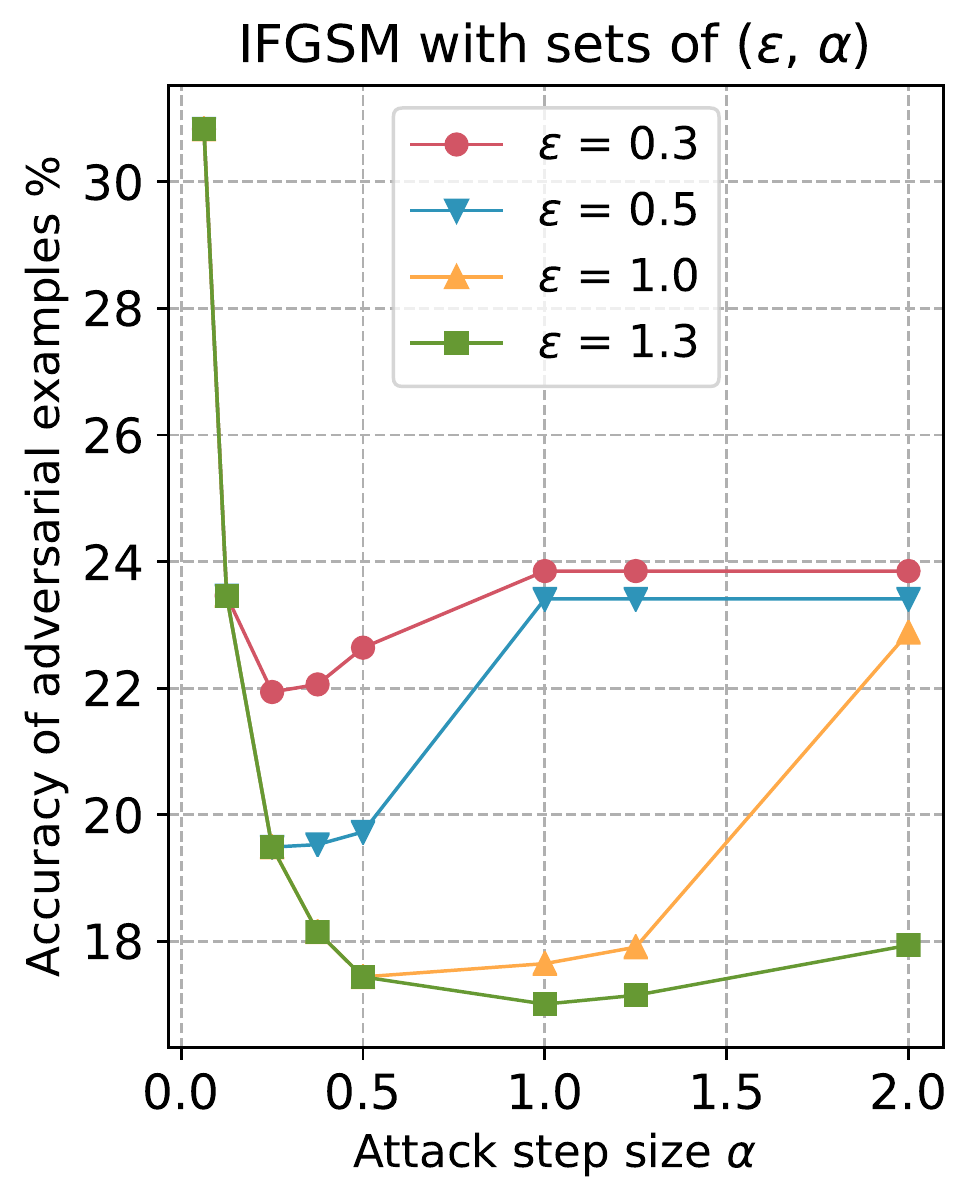}\label{ifgsme}
}%
\subfigure[]{
\centering
\includegraphics[width=1.2in]{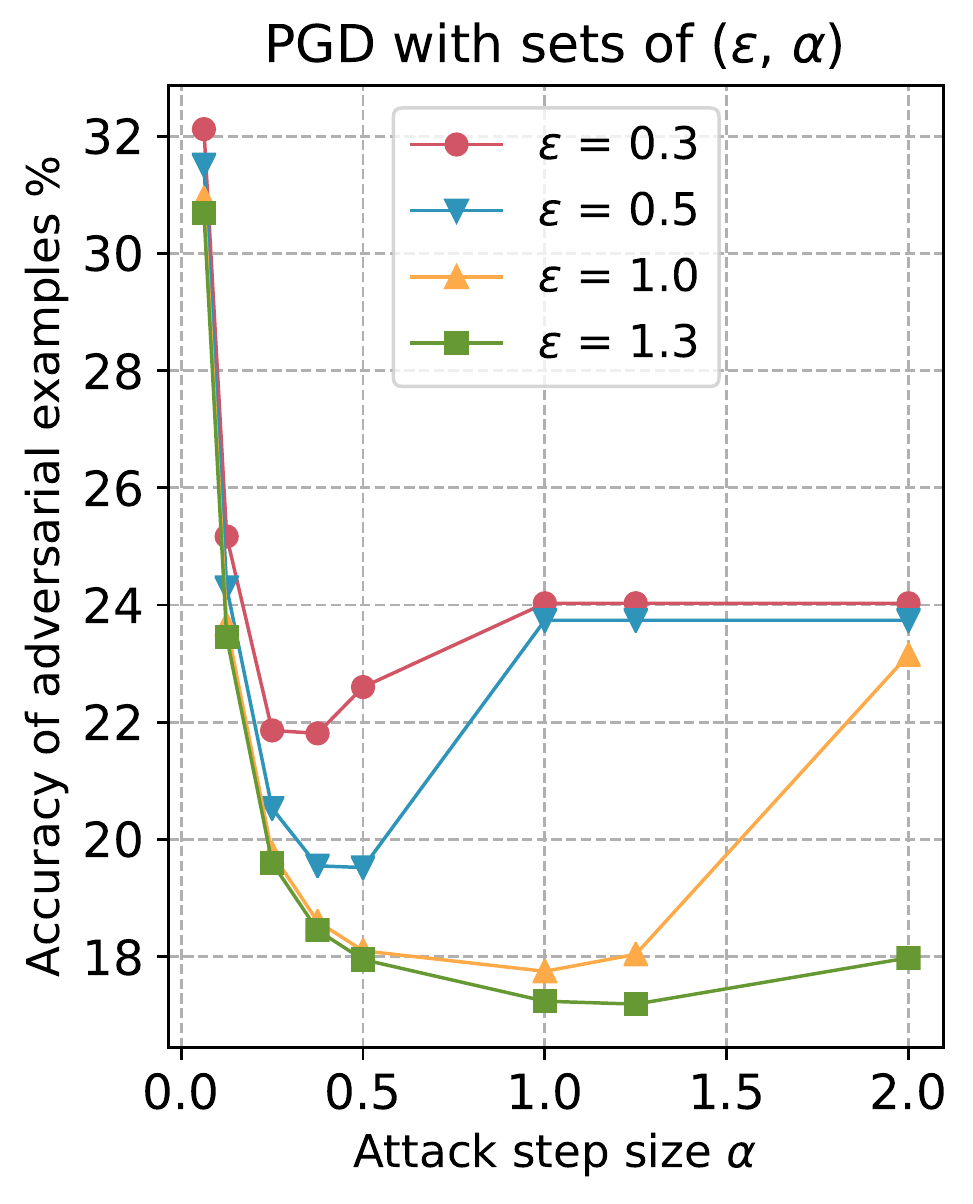}\label{pgde}
}%
\vspace{-3mm}
\quad
\vspace{-3mm}
\subfigure[]{
\centering
\includegraphics[width=1.22in]{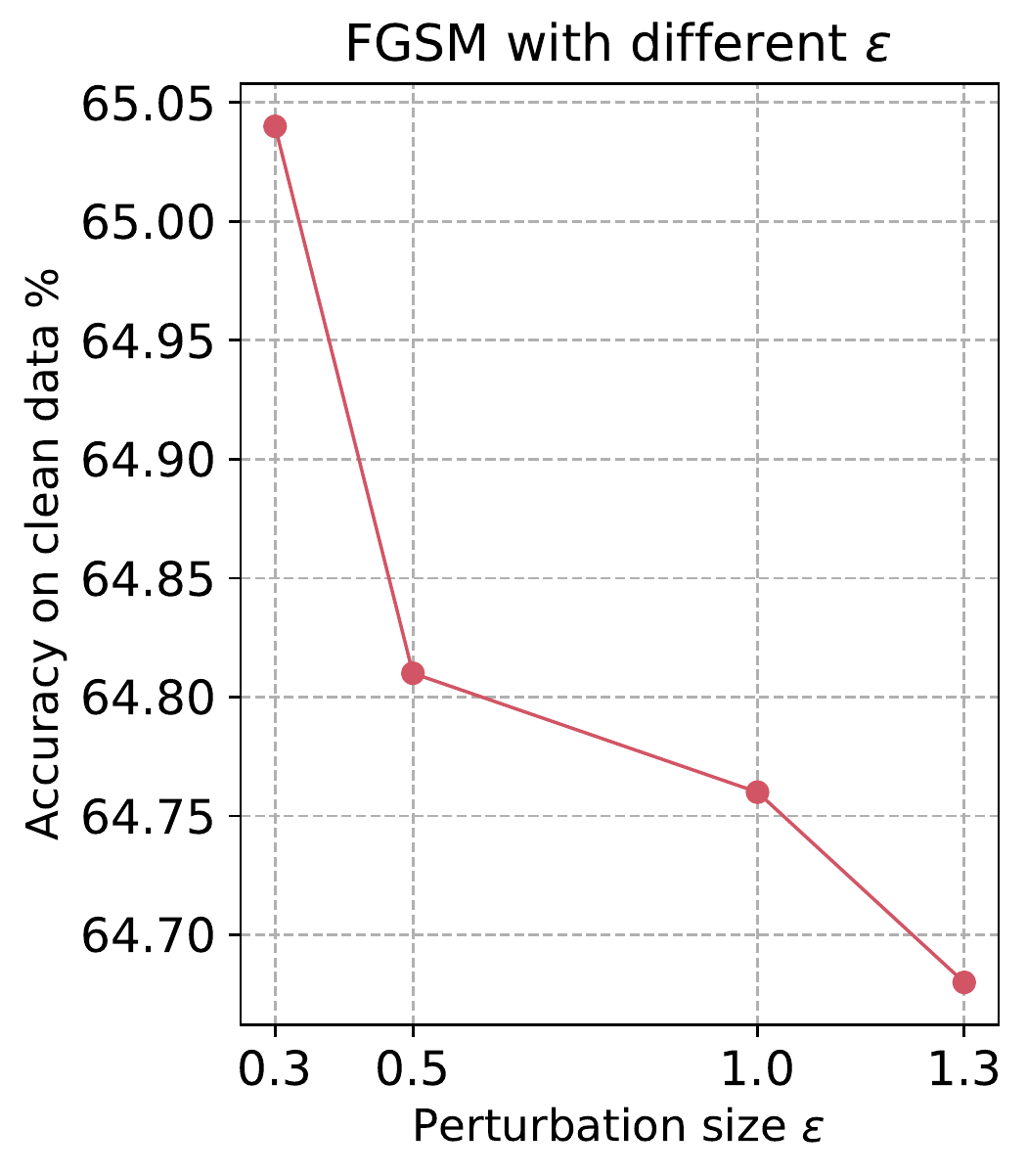}\label{fgsmp}
}%
\subfigure[]{
\centering
\includegraphics[width=1.2in]{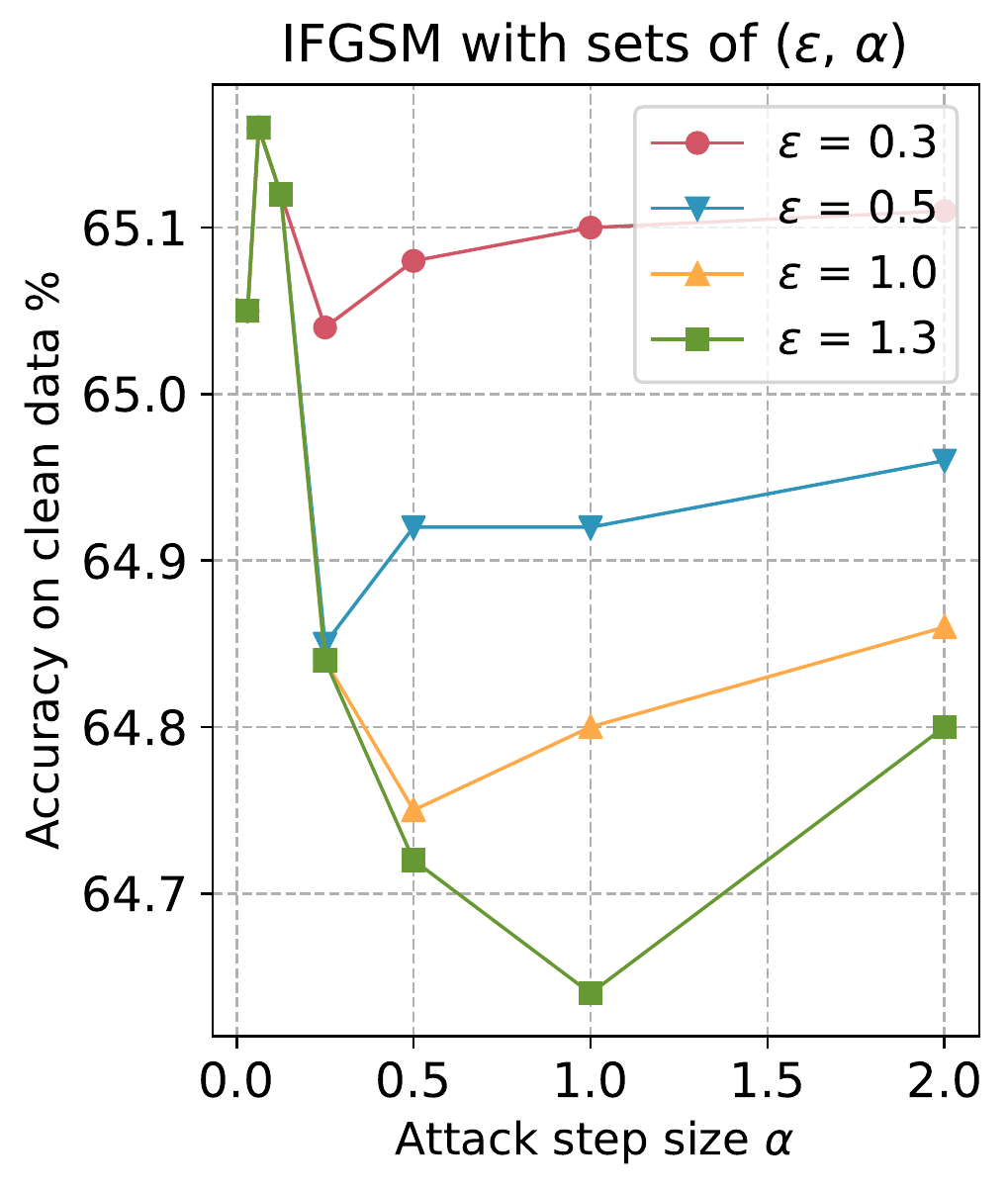}\label{ifgsmp}
}%
\subfigure[]{
\centering
\includegraphics[width=1.2in]{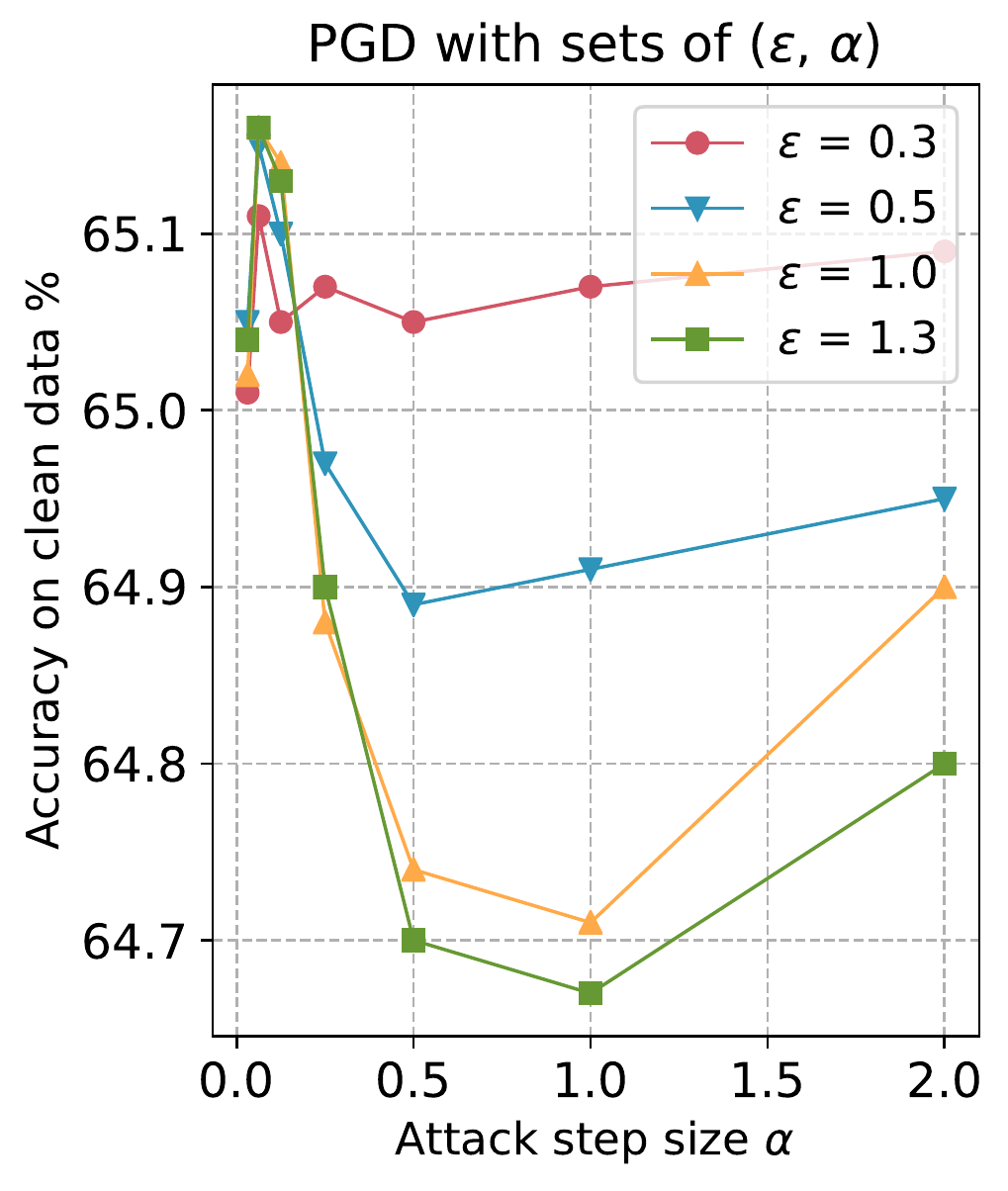}\label{pgdp}
}%
\centering
\caption{Ablation on visual attacker strength and type. 
 The top row is the accuracy of the vanilla model on adversarial examples generated by FGSM, IFGSM, and PGD, respectively. The bottom row is the standard accuracy of our model that adversarially trained with the corresponding attacker. The number of iterations is fixed to 2.}
\label{figbig}
\end{figure}
\vspace{-10pt}
\subsubsection{Effect of Augmenting Time.}
We empirically find that the time when the adversarial examples are injected into training has an effect on accuracy. We demonstrate this via ablation studies in Table \ref{epoch}. 
We try several adversarial training period $(5,25)$, $(10,25)$, $(10,15)$ $(15, 20)$. They respectively evaluate the effect of delaying injecting additional training data after different epochs and prove the advantage gained from fine-tuning with clean data in the last few epochs. Results show that $(10,15)$ is the optimal adversarial training period, and it surpasses the baseline model and achieves 65.16\% accuracy. One explanation is that adversarial examples have different underlying distributions to normal examples, and if boosting model performance on clean examples is our main goal, the fitting ability of model on clean examples need to be retrieved at the beginning and the end of the training process. It is inappropriate to inject the perturbed examples at an early stage where the model has not warm up.  
We fix the adversarial training period as $(10,15)$ and reuse the same partially trained (for 10 epochs) model as a starting point for the other ablation experiments.

\subsection{Ablation Studies}
\subsubsection{Augmentation Decomposed.} 
Results from ablation studies to test the contributions of our method's components are given in Table \ref{tab1}. The augmentation on visual and textual (question) data both make their individual contribution to improve the accuracy. We observe that visual adversarial examples are critical to our performance, and removing it causes a 0.47\% accuracy drop (see Ours $w/o$ Aug-V) on the validation set. The question augmentation also leads to comparable improvements, see the model of Ours $w/o$ Aug-V. 
\vspace{-10pt}
\subsubsection{Ablation on Adversarial Attackers.}
We now ablate the effects of attacker strength and type used in our method on network performance. 
To evaluate the regularization power of adversarial examples, we first compute the accuracy of the vanilla model after being attacked by the gradient-based attacker with a variety of sets of parameters. 
 Since the visual input ranges from 0 to 83, we try perturbation size $\epsilon$ among $\left\{0.3, 0.5, 1, 1.3 \right\}$, approximately following the ratio of $\epsilon$ to pixel value in \cite{advprop}, and step size $\alpha$ among $\left\{\frac{1}{16}, \frac{1}{8}, \frac{1}{4}, \frac{3}{8}, \frac{1}{2}, 1, 2 \right\}$. 

Fig. \ref{ifgsme} reflects the attacker strength changes with different parameter settings (accuracy declines implies strength increases) while Fig. \ref{ifgsmp} reflects how the model performance changes with attacker strength. Obviously, the accuracy decline when $\epsilon$ increases.
We observe that the accuracy on clean data is inversely proportional to attacker strength. As weaker attackers push the distribution of adversarial examples less away from the distribution of clean data, the model is better at bridging domain differences. However, the extremely weak attacker (e.g., random noise, $\alpha<\frac{1}{64}$) yields negligible improvement on standard accuracy, since the generated data are drawn similar distribution with original data.

We then study the effects of applying different gradient-based attackers in our method on model performance. Specifically, we try two other attackers, FGSM and PGD \cite{pgd}. FGSM is the one-step version of IFGSM, and PGD is a universal “first-order adversary” that adds the random noise initialization step to IFGSM. Their performances are reported in Fig. \ref{fgsme}, \ref{fgsmp}, \ref{pgde}, \ref{pgdp}. We observe that all attackers substantially improve model performance over the vanilla training baseline. The two iterative attackers obtain almost the same results while FGSM is less competitive. This result suggests that our VQA data augmentation is not designed for a specific attacker.

\subsection{Model Robustness}
Improvement of model robustness against adversarial attacks is a reward of our adversarial training scheme. 
As shown in Table \ref{big}, we are able to significantly increase accuracy on visual adversarial examples by 13.74\%, when using the training attacker at test-time. Following \cite{carlini2019evaluating}, we test a stronger PGD attacker ($\epsilon=0.5,\alpha=0.125,n=6$) and our model could beat the baseline by 4.59\%.
On the textual side, the accuracy of the vanilla model on $q_{adv}$ is 54.03\% and the flip rate (the rate of changing the original answers, lower is better) is 36.72\% while our adversarially trained model obtained an accuracy of 63.18\% and a flip rate of 18.8\% on $q_{adv}$. When attacking both visual and textual sides in test-time, our model beats the vanilla model by 21.55\%.
These results indicate that our model is capable of defending against both visual and textual common attackers.

\subsection{Human Evaluation of Semantic Consistency}
In order to show the semantic consistency of our generated paraphrases with original questions, we conduct a human study.
We sampled 100 questions and their paraphrases with top1 semantic similarity score defined in Eq. \ref{sscore}, and asked 4 human evaluators to assign labels (e.g., positive for similar or negative for not similar). We averaged the opinions of different evaluations for each query to get a positive score of 84\%. It demonstrates that the majority of paraphrases are similar to the originals.

\setlength{\tabcolsep}{8pt}
\begin{table}[t]
\begin{center}
\caption{Validation accuracy (\%) of vanilla and adversarially trained (using IFGSM $\epsilon=0.3,\alpha=0.0625,n=2$) network on clean and adversarial examples with various test-time attackers. Parap. represents the generated paraphrases in our method. Note that the IFGSM and PGD still act as the white-box attacker when testing.}
\label{big}
\begin{tabular}{lccccc}
\toprule
 & Clean & IFGSM & Parap. & IFGSM $\&$ Parap. &PGD \\
\midrule
BUTD \cite{butd} &63.32  & 30.83  & 54.03  & 22.09 & 18.05\\
\ \ +Ours   & $\mathbf{65.16}$ & 44.57  & 63.18 & 43.64 & 22.64 \\
\bottomrule
\end{tabular}
\end{center}
\vspace{-5mm}
\end{table}
\setlength{\tabcolsep}{1.4pt}

\section{Conclusions}
 In this paper, we propose to generate visual and textual adversarial examples as augmented data to train a robust VQA model with our adversarial training scheme. The visual adversaries are generated by gradient-based adversarial attacker and textual adversaries are paraphrases. Both of them keep modification imperceptible and maintain the semantics.
 Experimental results show that our method not only outperforms prior arts of VQA data augmentation, and also improves model robustness against adversarial attacks. To the best of our knowledge, this is the first work that uses both semantic equivalent visual and textual adversaries as data augmentation for the visual question answering problem.
 \vspace{-15pt}
 \subsubsection{Acknowledgements.}
This work was supported by National Key Research and Development Program of China (2016YFB1001003), NSFC (U19B2035, 
61527804, 60906119), STCSM (18DZ1112300). C. Ma was sponsored by Shanghai Pujiang Program.

%
%
\bibliographystyle{splncs04}
\bibliography{egbib}
\end{document}